%% file: multi14.tex
\documentclass[11pt]{article}

\usepackage{amsmath,amssymb}
\usepackage{amsthm}
\usepackage{mathtools}
\usepackage{fullpage}
\usepackage{makeidx}
\usepackage{enumerate}
\usepackage[top=1in, bottom=1.25in, left=1in, right=1in]{geometry}
\usepackage{graphicx,float,psfrag,epsfig,caption}
\usepackage[usenames,dvipsnames,svgnames,table]{xcolor}
\definecolor{darkgreen}{rgb}{0.0,0,0.9}
\usepackage[pagebackref,letterpaper=true,colorlinks=true,pdfpagemode=none,citecolor=OliveGreen,linkcolor=BrickRed,urlcolor=BrickRed,pdfstartview=FitH]{hyperref}
\usepackage{epstopdf}
\usepackage{color}
\usepackage{xr}
\usepackage{subfig}
\usepackage{caption}
\usepackage{graphicx}
\usepackage[utf8]{inputenc}
\usepackage{scalerel,stackengine}

\usepackage[boxed]{algorithm2e}
\usepackage[noend]{algorithmic}
\SetAlCapSkip{1em}

\DeclareMathAlphabet{\mathpzc}{OT1}{pzc}{m}{it}

\newtheorem{propo}{Proposition}[section]
\newtheorem{lemma}[propo]{Lemma}

\newtheorem{assumption}[propo]{Assumption}

\newtheorem{thm}[propo]{Theorem}

\addtocontents{toc}{\protect\setcounter{tocdepth}{2}}

\input{notation}

\title{Multi-Product Dynamic Pricing in High-Dimensions with Heterogeneous Price Sensitivity} 

\author{Adel~Javanmard \quad Hamid~Nazerzadeh \quad Simeng Shao\\ 
\\            
{{\small Data Sciences and Operations Department, University
of Southern California}}\\
{{\small Email: \{ajavanma,hamidnz,simengsh\}@usc.edu}}
}

\begin{document}

\maketitle

\begin{abstract}
We consider the problem of  multi-product dynamic pricing, in a contextual setting, for a seller of differentiated products. In this environment, the customers arrive over time and products are described by high-dimensional feature vectors. Each customer chooses a product  according to the widely used  Multinomial Logit (MNL)  choice model and her utility depends on the product features as well as the prices offered. 
The seller a-priori does not know the parameters of the choice model  but can learn them through interactions with customers. The seller's goal is to design a pricing policy that maximizes her cumulative revenue. 
 This model is motivated by online marketplaces such as Airbnb platform and online advertising. 
We measure the performance of a pricing policy in terms of regret, which is the expected revenue loss with respect to a clairvoyant policy that knows the parameters of the choice model in advance and always sets the revenue-maximizing prices.
We propose a pricing policy, named M3P, that achieves a $T$-period regret of $O(\log(Td) ( \sqrt{T}+ d\log(T)))$ under heterogeneous price sensitivity for products with features of dimension $d$. We also use tools from information theory to prove that no policy can achieve worst-case $T$-regret better than $\Omega(\sqrt{T})$.
\end{abstract}

\section{Introduction}
Online marketplaces offer very large number of products described by a large number of features. This contextual information creates differentiation among products and also affects the willingness-to-pay of buyers.  To provide more context, let us consider the Airbnb platform: the products sold in this market are ``stays." In booking a stay, the customer first selects the destination city, dates of visit, 
type of place (entire place, 1 bedroom, shared room, etc)  and hence narrows down her choice to a so-called \emph{consideration set}. The platform then sets the prices for the products in the consideration set. 
Notably, the products here are highly differentiable. Each product can be described by a high-dimensional feature vector that encodes its properties, such as  space, amenities, walking score, house rules, reviews of previous tenants, and so on. We study a model where the platform aims to maximize its cumulative revenue. 

In setting prices, there is a clear tradeoff: A high price may drive the customer away (decreases the likelihood of a sale) and hence hurts the revenue. A low price, on the other hand encourages the customer to purchase the product; however, it results in a smaller revenue from that sell. Therefore, in order for the seller to maximize her revenue, she needs to learn the purchase behavior of the customers through interactions with them and observing their purchasing decisions. Namely, the seller should learn how customers weigh different features in their purchasing decisions.


In this work, we study a setting where the utility from buying a product is a function of the product features and its price. 
\if false
Let $u({\theta}_0,\gamma_0,x,p)$ be the utility obtained from buying a product with feature vector ${x}$, at price $p$ where the parameter vectors ${\theta}_0$ and $\gamma_0$, in $\reals^d$, represent the customer's purchase behavior. Namely, $\theta_0$ captures the contribution of each feature to the customer's valuations of the products and $\gamma_0$ captures the sensitivity of the utility to the price. We focus on the utility model:
\begin{align}\label{eq:model0}
u({\theta}_0,{x},p) = \<x,{\theta}_0\> - \<x,\gamma_0\> p + z\,,
\end{align}
where  $\<{a},{b}\>$ indicates the inner product of two vectors ${a}$ and ${b}$. 
The term $z$, a.k.a. market noise, captures the idiosyncratic change in the valuation of each customer. Throughout, we call $\beta_x = \<x,\gamma_0\>$ the {\emph{price sensitivity}} parameter of a product with feature $x$. 
As we will discuss in Remark~\ref{rem2}, our utility model can be extended to models that are non-linear in the product feature $x$, and can capture several important classes such as log-log models, semi-log models as well as logistic models.
Also note that our utility model can recover the common case of ``value minus price" utility.
%
%
%
%
%

We encode the ``no-purchase" option as a new product with zero utility. 
We emphasize that \emph{the parameters of the utility model, ${\theta}_0$ and $\gamma_0$, are a priori unknown to the seller}. 

 \fi
In our model, given a consideration set, the customer chooses the products that results in the highest utility. We study the widely used Multinomial Logit (MNL) choice model~\cite{mcfadden1973conditional} 
\if false
which corresponds to having the noise term $z$ in Eq~\eqref{eq:model0} drawn independently from a standard Gumbel distribution, whose cdf is given by $G(y) = e^{-e^{-y}}$~\cite{talluri2006theory}.
The MNL model
\fi
which is arguably the most popular random utility model with a long history in economics and marketing research to model choice behavior~\cite{mcfadden1973conditional,mcfadden2001economic,guadagni2008logit}.  
We propose a dynamic pricing policy, called M3P,  for {\bf M}ulti-{\bf P}roduct {\bf P}ricing {\bf P}olicy in high-dimensional environments. Our policy uses maximum likelihood method to estimate the true parameters of the utility model based on previous purchasing behavior of the customers.
We measure the performance of a pricing policy  in terms of the regret, which is the difference between the expected revenue obtained by the pricing policy and the revenue gained by a clairvoyant policy that has full information of the parameters of the utility model and always offers the revenue-maximizing price. Our policy, achieves a $T$-regret of $O(\log(Td) ( \sqrt{T}+ d\log(T)))$, where $d$ and $T$ respectively denote  the features dimension and the length of the time horizon. Furthermore, we also prove that our policy is almost optimal in the sense that no policy can achieve worst-case $T$-regret better than $\Omega(\sqrt{T})$.

In the next section, we briefly review the related work to ours. We would like to highlight that our work is distinguished from the previous literature in two major aspects: (1) Multi-product pricing that should take into account the interaction of different products as changing the price for one product may shift the demand for other products, and this dependence makes the pricing problem even more complex. (2) Heterogeneity and uncertainty in price sensitivity parameters. 
\if false
There has been a recent line of work for dynamic pricing under contextual information with parametric learning that achieves $O_d(\log T)$ regret~\cite{cohen2016feature,javanmard2016dynamic,ban2017personalized,golrezaei2019dynamic,lobel2018multidimensional,leme2018contextual}. We will discuss them in the next section and we argue that the logarithmic regret of these policies do not carry over to our setting mainly due to point $(ii)$ above, that is the fact that the price sensitivity is unknown to the seller's policy in advance. 

We believe that is an important insight as it sheds light on the difference between the optimal regret bounds under different settings.   
\fi

\subsection{Related Work}
There is a vast literature on dynamic pricing as one of the central problems in revenue management. 
We refer the reader to~\cite{den2015dynamic,aviv2012dynamic} for extensive surveys on this area. 
A popular theme in this area is dynamic pricing with learning where there is uncertainty about the demand function, but information about it can be obtained via interaction with customers. A line of work~\cite{araman2009dynamic,farias2010dynamic,harrison2012bayesian,cesa2015regret,ferreira2016online,cheung2017dynamic} took Bayesian approach for the learning part and studied this problem in non-contextual setting.
Another related line of work assumes parametric models for the demand function with a small number of parameters, and proposes policies to learn these  parameters using statistical procedures such as maximum likelihood~\cite{besbes2009dynamic,broder2012dynamic,den2013simultaneously,debBoerZwart2014,chen2015statistical} or least square estimation~\cite{besbes2009dynamic,goldenshluger2013linear,keskin2014SCV}. 

Recently, there has been an interest in dynamic pricing in contextual setting. The work~\cite{amin2013learning,cohen2016feature,lobel2018multidimensional,javanmard2016dynamic,ban2017personalized} consider single-product setting where the seller receives a single  product at each step to sell (corresponding to $N=1$ in our setting) and assume equal price sensitivities $\beta=1$ for all products. In~\cite{amin2013learning}, the authors consider a noiseless valuation model with strategic buyer and propose a policy with $T$-period regret of order $O(T^{2/3})$. This setting has been extended to include market noise and also a market of strategic buyers who are utility maximizers~\cite{golrezaei2019dynamic}. In~\cite{cohen2016feature}, authors propose a pricing policy based on binary search in high-dimension with adversarial features that achieves regret $O(d^2 \log (T/d))$. This was later improved to $O(d\log T)$ in \cite{lobel2018multidimensional}.
Using ideas from integral geometry,~\cite{leme2018contextual} proposed a  contextual decision-making policy using binary observations that achieves $O_d(\log \log T)$ regret, and also extended similar ideas to a more general setting of learning Lipschitz functions from binary feedbacks~\cite{mao2018contextual}. 
The work~\cite{javanmard2016dynamic} studies the dynamic pricing in high-dimensional contextual setting with sparsity structure and propose a policy with regret $O(s_0 \log (d) \log (T))$ but, again in a single-product scenario and unit price sensitivity. The dynamic pricing problem has also been studied under time-varying coefficient valuation models~\cite{javanmard2017perishability} to address the time-varying purchase behavior of buyers and the perishability of sales data. 

Let us emphasize again that our setting deviates from the settings studied in these paper in two main directions: 
\begin{enumerate}
\item We are considering multi-product case, where at each round a \emph{pool} of (varying) products are offered to the buyers. This allows us to better model some important applications, such as the pricing problem for Airbnb platform.
\item We consider heterogeneous price sensitivities for the products, and the price sensitivities are unknown to the seller.
\end{enumerate}
Specifically, due to point (2) above the logarithmic regret bounds established in the previous work, discussed above, will not apply to our setting and we indeed prove $\Omega(\sqrt{T})$ lower bound for the worst-case $T$- period regret of any pricing policy in our setting. Our derivation of the lower bound, cf. Theorem~\ref{thm:M3P-2}, is by constructing special instances of the MNL model for which there is a tradeoff between learning (reducing uncertainty about) the model parameters and exploiting the best-guess optimal prices. We use KL-divergence as a quantitive measure of uncertainty and study the tradeoff between KL-divergence and the regret of a policy which results in the $\Omega(\sqrt{T})$ lower bound. 

Very recently,~\cite{mueller2018low} studied high-dimensional multi-product pricing, with a low-dimensional linear model for the aggregate demand. In this model, the demand vector for all the products at each step is observed, while in our work the seller only sees the product index that is chosen by the buyer at each step. Similarly,~\cite{qiang2016dynamic} studies
a model where the seller can observe the aggregate demand and proposes a myopic policy based on least-square estimations that obtains a
logarithmic regret.

Finally, note that the problem studied in this paper is a  contextual bandit problem with a specific structure, revealed in the reward function. Contextual bandit problems are well studied in the literature. We omit a review of this area in the interest of space and refer to~\cite{auer2002mutiarmed,langford2008epoch,agarwal2012contextual}.

\subsection{Contributions and Challenges} This work contributes to the literature of dynamic pricing by
1) problem formulation and modeling that captures the pricing problem faced by many online marketplaces, such as Airbnb platform; 2) proposing a novel pricing policy that uses the contextual information effectively and achieves a low regret with respect to both time and feature dimension;  3) analysis of the proposed policy and establishing a lower bound on the regret of any pricing policy. Our analysis also provides important insights on the role of price sensitivity and its uncertainty to the seller in the best achievable regret. Indeed our $\Omega(\sqrt{T})$ lower bound demonstrates that some of the recent results in dynamic pricing with parametric learning,  achieving $O_d(\log T)$ regret, do not carry over to our setting due to heterogeneous price sensitivity. This insight helps with understanding the applicability of various pricing policies and their regret bounds.  

In our analysis we derive bounds on the performance of maximum likelihood (ML) estimator that is used to learn the model parameters. It is worth noting that the seller's observation at each step (choice made by the buyer) depends on seller's action (posted price), which in turn depends on the previous sales data.  That said, the samples used in the ML estimator are correlated and hence classical results in parametric regression do not carry over to our setting. Even more, looking at the price term $\<x_i,\gamma_0\> p_{it}$ in the utility, one can think of $w_{it} \equiv p_{it}x_i$ as new regressor in measuring $\gamma_0$. However,
 common methods and analyses for the estimation error requires the features to be well separated (so that
 each measurement gives new information about the underlying parameter of interest.) But the features $w_{it}$ can be very correlated via the prices (especially for policies that choose prices in a way to greatly exploit the previous sales data). These are among the technical challenges that have been addressed in our analysis.

\section{Model}  \label{sec:model}
We consider a firm which sells a set of products to customers that arrive over time. The products are differentiated and each is described by a wide range of features.
At each step $t$, the customer selects a consideration set $\cC_t$ of size at most $N$ from the available products. This is the set the customer will actively consider in her purchase decision.  The seller sets the price for each of the products in this set, after which the customer may choose (at most) one of the products in $\cC_t$. If he chooses a product, a sale occurs and the seller collects a revenue in the amount of the posted price; otherwise, no sale occurs and seller does not get any revenue.


Each product $i$ is represented by an observable vector of features $x_i\in \reals^{{d}}$. Products offered at different rounds can be highly differentiated  and we assume that the feature vectors are sampled independently from a fixed, but unknown, distribution $\mathcal{D}\subset \reals^{{d}}$. For the sake of normalization, we assume that the support of $\mathcal{D}$ is a subset of $[-1,1]^d$. 


If an item $i$ (at period $t$) is priced at $p_{it}$, then the customer obtains utility $u_{it}$ from buying it, where\footnote{In general the offered price not only depends on the feature vectors ${x}_i$ but also the period $t$, as the estimate of the model parameters may vary across time $t$. We make this explicit in the notation $p_{it}$ by considering both $i$ and $t$ in the subscript.} 
\begin{align}\label{eq:utility-model-linear}
u_{it} = \<x_i,\theta_0\> - \<x_i,\gamma_0\> p_{it} +z_{it}\,.
\end{align} 
Here, $\theta_0, \gamma_0 \in \reals^d$ are the parameters of the demand curve and are \emph{unknown} a priori to the seller.  We assume $\|({\theta}_0,\gamma_0)\| \le \l1u$, for an arbitrarily large but fixed constant $\l1u$, with $\|\cdot\|$ indicating the $\ell_2$ norm.
Note that this is a random utility model with $z_{it}$ component representing market shocks (noise). 

At each step the user chooses the item with maximum utility from her consideration set; in case of equal utilities, we break the tie randomly.
	\medskip
	
	To summarize, our setting is as follows. At each period $t$:
	\begin{enumerate}
		\item The customer narrows down her options by forming a consideration set $\cC_t$ of size at most $N$.
		\item For each product $i\in \cC_t$, the seller offers a price $p_{it}$.\footnote{Equivalently, the seller can determine all the prices in advance and reveal them  after the customer determines the consideration set. We note that the consideration set of the customer does not depend on the prices, but the choice she makes from the consideration set depends on the prices. In addition, recall that {all} the customers share the same $\theta_0$ and $\gamma_0$ and the choice of consideration set does not reveal information about these parameters.}
		\item The customer chooses item $i_t\in \cC_t{\cup\{\emptyset\}}$ where $i_t = \arg\max_{i\in \cC_t{\cup\{\emptyset\}}} u_{it}$. 
		\item The seller observes the product chosen from the consideration set and uses this information to set the future prices. 
	\end{enumerate}

In this work, we consider the multinomial logit (MNL) choice model that has been widely used by practitioners and researchers (arguably the most popular random utility model, which is derived from the \emph{``independence of irrelevant alternative"} axiom as a choice model given a pool of options). It has quite a long history
in economic and transportation research \cite{mcfadden2001economic}, as well as marketing \cite{hauser1979alternative,gensch1979multinomial,louviere1983design,guadagni2008logit}.
Positing the MNL model is \emph{equivalent} to assuming the market shocks $z_{it}$ being drawn independently and identically from the standard Gumbel distribution. Under the MNL model, the probability of choosing an item $i$ from set $\cC_t$ is given by
\begin{eqnarray}\label{probs}
q_{it} \equiv \prob(i_t = i | \cC_t) = \frac{\exp(u^0_{it})}{1+ \sum_{\ell\in \cC_t} \exp(u^0_{\ell t})}\,, \text{ for } i\in \cC_t\,,
\end{eqnarray}
where $u^0_{it} = \<x_i,\theta_0\> -\<x_i,\gamma_0\> p_{it}$, for $i\in \cC_t$.

We refer to the term $\beta_{i} = \<x_i,\gamma_0\>$ in the utility model as the {price sensitivity} of product $i$. Note that our model allows for heterogeneous price sensitivities. We also encode the no-purchase option by item $\emptyset$, with market utility $z_{\emptyset t}$, drawn from zero mean Gumbel distribution. The random utility $z_{\emptyset t}$ can be interpreted as the utility obtained from choosing an option outside the offered ones. This is equivalent to $u^0_{\emptyset t} = 0$. 

%

We make the following assumption that ensures positivity of the products price sensitivity parameters. Per this assumption, note that for a product with feature $x_i\in \mathcal{D}$, the price sensitivity is given by $\<x_i, \gamma_0\>$.  
\begin{assumption}\label{ass1}
We have $\min \{\<x,\gamma_0\>:\, {x\in \mathcal{D}}\} \ge  L_0 > 0$, for some constant $L_0$.
\end{assumption} 
\if false
\begin{remark}\label{rem2}
Although we focus on linear utility model~\eqref{eq:utility-model-linear}, it is  straightforward to extend our pricing policy and its analysis to some nonlinear utility models. Specifically, consider the utility function 
\[
u_{it} = \psi\Big(\<\phi(x_i),\theta_0\> - \<\phi(x_i),\gamma_0\> + z_{it} \Big)\,,
\]
where $\phi$ is a feature mapping and $\psi:\reals\mapsto \reals$ is a general strictly increasing function. 
Examples of such utility models include: $(i)$ Log-log model ($\psi(y) = e^{y}$, $\phi(y) = \ln(y)$); $(ii)$ Semi-log model ($\psi(y) = e^{y}$, $\phi(y)= y$); and $(iii)$ Logistic model ($\psi(y) = e^{y}/(1+e^{y})$, $\phi(y) = y$).

Note that by the change of variable $\tilde{u}_{it} = \psi^{-1}(u_{it})$, $\tx_i = \phi(x_i)$, we get back to a utility model of form $\tilde{u}_{it} = \<\tx_i, \theta_0\> - \<\tx_i,\gamma_0\> + z_{it}$, which is linear in the mapped feature $\tx_i$. Given the monotone-increasing assumption of $\psi$, the product with the highest utility $u_{it}$ would be the same with the highest utility $u_{it}$ and in this sense the choice model is unaltered under this change of variable. As a result, we only need to apply this change of variable in the proposed M3P to address the nonlinear utility model. The analysis also carries over readily at gives a $T$- period regret of $O(\log(Td)(\sqrt{T}+ d\log(T)))$.
\end{remark}
\fi
%

Before proceeding with the policy description, we will discuss the benchmark policy which is used in defining the notion of regret and measuring the performance of pricing policies.  

\section{Benchmark policy}
The seller's goal is to minimize her regret, which is defined as the expected revenue loss against a clairvoyant policy that knows the utility model parameters $\theta_0, \gamma_0$ in advance and always offers the revenue-maximizing prices. Formally, let $\pi$ be a pricing policy that chooses prices $p^\pi_t = (p^\pi_{it})_{i\in \cC_t}$ at time $t$ for the products in the consideration set $\cC_t$. Then, the seller's expected revenue at period $t$, under such policy will be 
	\begin{align}\label{rev}
	\rev^\pi_t = \sum_{i\in \cC_t} q_{it} p^\pi_{it}\,,
	\end{align} 
	with $q_{it}$ being the probability of buying product $i$ from the set $\cC_t$ as given by Eq~\eqref{probs}.\footnote{More precisely, $\rev^\pi_t$ is the expected revenue conditional on filtration $\cF_{t-1}$, where $\cF_t$ is the sigma algebra generated by feature matrices $X_1, \dotsc, X_{t+1}$ and market shocks $z_1,\dotsc, z_{t}$.} Similarly, we let $\rev^*_t$ be the seller's expected revenue under the benchmark policy that sets price vectors {{${p}^*_t$}}, at period $t$. 
	The worst-case cumulative regret of policy $\pi$ is defined as
	\begin{align*}
		\Regret^\pi(T) \equiv \sup_{\|({\theta}_0,\gamma_0)\|\le W} \quad\sum_{t=1}^T (\rev^*_t - \rev^\pi_t)\,.
		\end{align*}
We next characterize the benchmark policy. Let $p^*_t = (p^{*}_{it})_{i\in \cC_t}$ and $X_t\in \reals^{|\cC_t| \times d}$ be the feature matrix, which is obtained by stacking $x_i$, $i\in \cC_t$ as its rows (Recall that $|\cC_t| \le N$). The proposition below gives an implicit formula to write the vector of optimal prices $p^*_t$ as a function $p^*_t = g(X_t\gamma_0, {X}_t\theta_0)$. We refer to $g$ as the pricing function.
\begin{propo}\label{propo:opt-price}
The benchmark policy that knows the utility model parameters $\theta_0, \gamma_0$, sets the optimal prices as follows. For product $i\in \cC_t$, the optimal price is given by 
\begin{align}\label{eq:function-g}
p^*_{it} = \frac{1}{\<x_i,\gamma_0\>}+ B^{0}_{t} \equiv (g(X_t\gamma_0, {X}_t\theta_0))_i\,,
\end{align}
where $B^{0}_{t}$ is the unique fixed point $B$ of the following equation:
\begin{align}\label{eq:opt-price}
B = \sum_{\ell\in \cC_t} \frac{1}{\<x_{\ell},\gamma_0\>} e^{-(1+\<x_{\ell},\gamma_0\> B)} e^{\<x_{\ell},\theta_0\>}\,. 
\end{align}
\end{propo}
The proof follows by writing the first order optimality of the seller's revenue as a function of prices and rearranging the terms.
The uniqueness of the solution follows from the fact that in \eqref{eq:opt-price} the left hand side is strictly increasing in $B$ and is zero at $B=0$, while the right hand side is strictly decreasing in $B$ and is positive at $B=0$. 
We refer to the appendix for the proof of Proposition~\ref{propo:opt-price}. 
\begin{algorithm}[t]
\caption{M3P policy for multi-product dynamic pricing}

\textbf{Input:\ (at time 0)} function $g$, parameter $W$ (bound on $\|({\theta}_0,\gamma_0)\|$)\\
\textbf{Input:\ (arrives over time)}
covariate matrices $\{{X}_t\}_{t\in [T]}$\\

\textbf{Output:\ prices $\{{p}_t\}_{t\in [T]}$}\\
1:\ $\tau_1 \leftarrow 1$, ${p}_1 \leftarrow {0}$, $\theta^{1} \leftarrow {0}$ and set the length of $k$-th episode: $\ell_k \leftarrow k+d$\\
2:\ \textbf{for} each episode $k = 1, 3, ...$ \textbf{do}\\
3:\quad \underline{Exploration Phase:} for the initial $d$ periods of episode $k$, choose the prices of items independently as $p_{it}\sim {\sf Uniform}([0,1])$.\\
4:\quad \underline{Exploitation Phase (learning):} At the end of the exploration phase, update the model parameter estimate $\hpsi^{k}$ using the ML estimator applied to the previous exploration periods:
\begin{align}\label{loss-opt-unkown}
\hpsi^{k}=\arg \min_{\| {\nu}\| \leq W} \mathcal{L}_k({\nu})\,,
\end{align}
with $\mathcal{L}_k({\nu})$ given by \eqref{eq:loss}.
%
\\
5:\quad \underline{Exploitation Phase (pricing):} offer prices based on the current estimate $\hpsi^{k}={\begin{bmatrix}
\hth^k\\
\hgamma^k
\end{bmatrix}}$ as
\begin{align}\label{tbl:price1}
    p_{it} \leftarrow \frac{1}{\<x_{it}, \hgamma^k\>} + B_t\,,
\end{align}
where $B_{t}$ is the unique value of $B$ satisfying  the following equation:
\begin{align}\label{tbl:price2}
B = \sum_{\ell\in \cC_t} \frac{1}{\<x_{\ell }, \hgamma^k\>} e^{-(1+\<\boldsymbol{x}_{\ell}, \hgamma^k\> B)} e^{\<\boldsymbol{x}_{\ell},\hth^{k}\>}\,. 
\end{align}
\end{algorithm}


\section{Multi-Product Pricing Policy (M3P)}

Here we provide a formal description of our multi-product dynamic pricing policy (M3P). The policy sees the time horizon in an episodic structure, where the length of episodes grow linearly . Specifically, episode $k$ is of length $\ell_k = k+d$, with the first $d$ referred to as exploration periods and the other $k$ ones as exploitation periods. Throughout, we use notation $E_k$ to refer to periods in episode $k$, i.e., $E_k = \{\ell_k,...,\ell_{k+1}-1\}$. 
In the exploration periods, the products prices are chosen independently as $p_{it} \sim {\sf Uniform}([0,1])$.\footnote{Indeed we can offer $p_{it} \sim {\sf Uniform}([0,C])$ for any fixed constant $C>0$ and $C$ appears in the regret bound. But since we treat $C$ as a constant it does not affect the order of the regret.} In the exploitation periods, we choose the optimal prices based on the current estimate of the model parameters. Concretely, let $I_k$ be the set of exploration periods up to episode $k$, i.e., the set consisting of the initial $d$ periods in episode $1, \dotsc, k$, and hence $|I_k| = kd$. We form the negative log-likelihood function for estimating $\nu_0 = (\theta_0^T, \gamma_0^T)^T$ using the sales data in periods $I_k$:
\begin{equation}\label{eq:loss0}
{\mathcal{L}_k({\nu}) =-\frac{1}{kd}{\displaystyle \sum_{t\in {I}_{k}}\log\frac{\exp(u^0_{i_tt}(\nu))}{\sum_{\ell\in\mathcal{C}_t\cup\{\emptyset \}}\exp(u^0_{\ell t}(\nu))}}\nonumber\,,}
\end{equation}
where $i_t$ denotes the product purchased at time $t$, and 
\begin{align}\label{utility0}
u^0_{it}(\nu) = \<{x}_i,\theta\> - \<x_i,\gamma\> p_{it}\,,
\end{align}
with {{$\nu = (\theta^{T}, \gamma^{T})^{T}$}}. We adopt the convention that $i_t = \emptyset$ for the ``no-purchase" case with $u^0_{\emptyset t}(\cdot)=0$.

The log-likelihood loss can be written in a more compact form. We let ${y}_t = (y_{it})_{i\in \cC_t}$ be the response vector that indicates which product is purchased at time $t$:
\[
y_{it}=\begin{cases}
1 & \text{product } i \text{ is chosen\,,}\\
0 & \text{otherwise\,.}
\end{cases}
\]
We also let ${u}^0_t (\nu) = (u^0_{it}(\nu))_{i\in \cC_t\cup \{0\}}$. Then, the log-likelihood loss can be written as
\begin{align}\label{eq:loss}
\mathcal{L}_k(\nu) =-\frac{1}{kd}{\displaystyle \sum_{t\in I_k}\log\frac{{y}_t \cdot \exp({u}^0_{t}(\nu))}{1+\sum_{\ell\in \mathcal{C}_t}\exp(u^0_{\ell t}(\nu))}}\,.
\end{align}
\begin{figure*}[h]
{{\centering
\includegraphics[scale = 0.9]{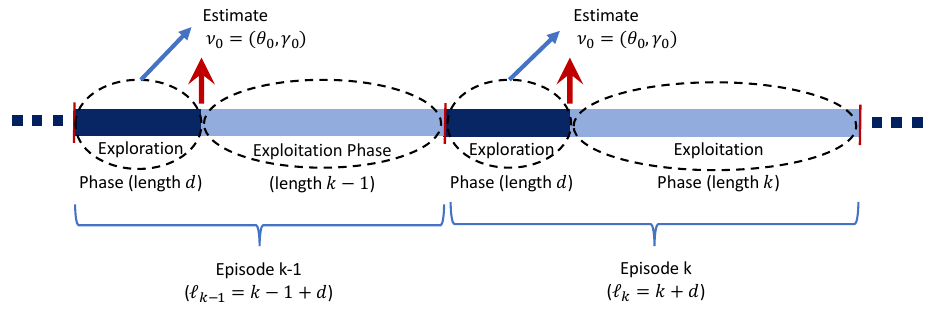}
 \caption{{\small Schematic representation of the Multi-product pricing policy (M3P). 
  It has an episodic structure, where each episode $k$ starts with an exploration phase of length $d$, followed by an exploitation phase of length $k$. M3P updates its estimates of the choice model parameters at the beginning of each exploitation phase. The estimates are computed via the maximum log-likelihood {method} using the buyer's choices from the previous exploration phase. The dark blue rectangles show the random exploration periods. }}\label{fig:corp}
}}
\end{figure*}
We construct the estimate $\hpsi^k = (\htheta^k, \hgamma^k)$ by solving the following optimization:
	\begin{align}\label{loss2}
	\hpsi^{k}=\arg \min_{\|\nu\| \leq W}\mathcal{L}_k(\nu)\,.
	\end{align}


In the exploitation periods of the $k$-th episode, the policy adheres to the estimate $\hpsi^k  = (\htheta^k, \hgamma^k)$ throughout the episode and sets the price vectors as $p_t = g(X_t \hgamma_t, X_t \hth^k)$, which is implicitly characterized in Eq~\eqref{eq:function-g}.
We refer to Figure~\ref{fig:corp} for a schematic representation of M3P.

The policy terminates at time $T$ but note that the policy does not need to know $T$ in advance. 
%
Also, by the design when the policy does not have much information about the model parameters it updates its estimates frequently (since the length of episodes are small) but as time proceeds the policy gathers more information about the parameters and updates its estimates less frequently, and use them over longer episodes.  

\section{Regret Analysis for M3P}\label{sec:regret-M3P}
We next state our result on the regret of M3P policy.
\begin{thm}{\bf (Regret upper bound)}\label{thm:M3P}
Consider the choice model~\eqref{probs}. Then, the worst case $T$-period regret of the M3P policy is of $O(\log(Td) ( \sqrt{T}+ d\log(T)))$, with $d$ and $T$ being the feature dimension and the length of time horizon.
\end{thm}

Below, we state the key lemmas in the proof of Theorem~\ref{thm:M3P} and refer to the appendix for the proof of technical steps. Let ${p}_t = (p_{it})_{t\in \cC_t}$ be the vector of prices posted at time $t$ for products in the consideration set $\cC_t$. 
Recall that M3P sets the prices as ${p}_t = g(X_t \hgamma^k, X_t \htheta^k)$, where $g(\cdot,\cdot)$ is the pricing function whose implicit characterization is given by Proposition~\ref{propo:opt-price}.

Our next lemma shows that the pricing function $g(\cdot,\cdot)$ is Lipschitz. We remind that $L_0$ is given in Assumption~\ref{ass1}, $W$ is the initial bound on the model parameters ($\|(\theta_0, \gamma_0)\| \le W$ as described in Section~\ref{sec:model}) and $N$ is the maximum size of the consideration set at each step. 
\begin{lemma}\label{lemma:price-lipschitz}
Suppose that $p_1 = g(X_t \gamma_1, X_t \theta_1)$ and $p_2 = g(X_t \gamma_2, X_t \theta_2)$. Then, there exists a
constant $C = C(W,L_0)>0$ such that the following holds
\begin{align}\label{eq:bound-p}
\| p_1- p_2\| \le C N^{2}\left (\|X_t(\gamma_1-\gamma_2)\|^2 + \|X_t(\theta_1-\theta_2)\|^2 \right)^{1/2}\,.
\end{align}
\end{lemma}
We next upper bound the right-hand side of Eq~\eqref{eq:bound-p} by bounding the estimation error of the proposed estimator. 
\begin{propo}\label{estimation-error}
Let $\hpsi^k$ be the solution of optimization problem~\eqref{loss2}. 
Then, there exist constants $c_0, c_1 $, and $c_2$ (depending on $W, L_0, N$), such that for $k\ge c_0d$,  with probability at least $1-d^{-2} k^{-1.5} -2e^{-c_2 kd}$, we have
\begin{align}\label{eq: estimation-error-bound}
\|\hth^k-{\theta}_0\|^2 + \|\hgamma^k-{\gamma}_0\|^2 &\leq {c_1 } \frac{\log(kd)}{k}\,, 
\end{align}
\end{propo}

The last part of the proof is to relate the regret of the policy at each period $t$ to the distance between the posted price vector ${p}_t$ and the price vector ${p}^*_t$ posted by the benchmark. Recall the definition of revenue $\rev^\pi_t$ from~\eqref{rev} and define the regret as $\mreg_t \equiv \rev^*_t - \rev^\pi_t$. 

\begin{lemma}\label{lemma:regret-lipschitz}
Let ${p}^*_t = g(X_t\gamma_0, X_t {\theta}_0)$ be the optimal price vector posted by the benchmark policy that knows the model parameters $\theta_0$ and $\gamma_0$ in advance.  There exists a constant $C>0$ (depending on $W$) such that the following holds,
\[
\mreg_t  \leq c_3 N \|p^{*}_{t} -p_{t}\|^{2}\,,
\]
for some constant $c_3 = c_3(W,L_0)$.
\end{lemma}
The proof of Theorem~\ref{thm:M3P} follows by combining Lemma~\ref{lemma:price-lipschitz}, Proposition~\ref{estimation-error} and Lemma~\ref{lemma:regret-lipschitz}. We refer to the appendix for its proof.

Our next theorem provides a lower bound on the $T$-regret of any pricing policy. 
\begin{thm}{\bf (Regret lower bound)}\label{thm:M3P-2}
Consider the choice model~\eqref{probs}. Then, the $T$-period regret of any pricing policy in this case is $\Omega(\sqrt{T})$.
\end{thm}

Theorem~\ref{thm:M3P-2} implies that  M3P has optimal cumulative regret in $T$, up to logarithmic factor.

\subsection{Proof Sketch for Theorem~\ref{thm:M3P-2}} To derive the lower bound, we pinpoint a tradeoff between reducing uncertainty about the model parameters $(\theta_0,\gamma_0)$, and exploiting the best-guess optimal price. Specifically, for any pricing policy $\pi$ and parameter $\theta\in {{\sf \Theta}}$ we let $f_t^{\pi,\theta}: \{0,1\}^t \to [0,1]$ be the probability distribution of the customer purchase responses $\by_t = (y_1, \dotsc, y_t)$ under pricing policy $\pi$ and model parameter $\theta$:
\[
f_t^{\pi,\theta}(\by_t) = \prod_{i=1}^t q(p_i,\theta)^{y_i} (1- q(p_i,\theta))^{1-{y_i}}\,,
\] 
where $p_i = \pi(\by_{i-1})$ is the price posted under policy $\pi$. In our analysis, we use the KL-divergence as a quantitative measure of uncertainty, that is to say a pricing policy $\pi$ has a large degree of certainty that the true parameter is $\theta_0$, rather than some counterfactual parameter $\theta$, if $\KL(f_t^{\pi,\theta_0}; f_t^{\pi,\theta})$ is large. Our Lemma E.2 in the appendix provides a lower bound of the form
\begin{align}\label{eq:dum-isit1}
\Regret^\pi(t,\theta_0) \gtrsim \frac{1}{(\theta_0-\theta)^2}\KL(f_t^{\pi,\theta_0}; f_t^{\pi,\theta})\,.
\end{align}
In words, reducing uncertainty about the model parameter is costly. In Lemma E.3 in the appendix, we prove the complementary part, showing that a policy that does not reduce its uncertainty about the model parameter also incur a regret cost. Namely, for model parameters $\theta_0$, $\theta_1 = \theta_0+ \frac{1}{4}T^{-1/4}$, 
\begin{align}\label{eq:dum-isit2}
\Regret^\pi(T,\theta_0) + \Regret^\pi(T,\theta_1) \gtrsim \sqrt{T} e^{-\KL(f_T^{\pi,\theta_0};f_T^{\pi,\theta_1})}\,.
\end{align}
The proof of~\eqref{eq:dum-isit2} relies on standard results on the minimum error probability of a two-hypothesis test. 

The lower bound $\Omega(\sqrt{T})$ is proved by combining bounds~\eqref{eq:dum-isit1} and \eqref{eq:dum-isit2} and using the inequality $u+e^{-u}\ge 1$ for all $u>0$.

We refer to the appendix for the detailed proof of Theorem~\ref{thm:M3P-2}.
\smallskip

\subsection*{Acknowledgements} This work was supported in part by a Google Faculty Research Award and the NSF CAREER Award
DMS-1844481.

{\hypersetup{linkcolor=black}

}

\bibliographystyle{abbrv}
 \bibliography{robust_learning_bib}

\newpage

\appendix
\section{Proof of Main Theorems}
\subsection{Proof of Theorem~\ref{thm:M3P}}\label{proof:thm-M3P}

By Lemma~\ref{lemma:regret-lipschitz}, we have
\begin{align}
\mreg_t 
\le c_3 N \|p_t-p_t^*\|^2 
&\le c_3 C^2 N^4 \left (\|X_t(\hgamma^k-\gamma_0)\|^2 + \|X_t(\htheta^k-\theta_0)\|^2 \right) \,, \label{reg1}
\end{align}
where we used Lemma~\ref{lemma:price-lipschitz}. 
Note that the estimates $\hgamma^k$ and $\htheta^k$ are constructed using the samples in $I_k$ and consequently are independent from the current features $X_t$ ($t$ is in the exploitation phase of episode $k$). Taking the expectation first with respect to $X_t$, we obtain
\begin{align}
\E(\mreg_t) &= c_3 C^2 N^5  \E\left [\|\Sigma^{1/2} (\hgamma^k-\gamma_0)\|^2 + \|\Sigma^{1/2} (\htheta^k-\theta_0)\|^2 \right] \nonumber\\
&\le c_3 c_{\max} C^2 N^5  \E\left[\|\hgamma^k-\gamma_0\|^2 + \| \htheta^k-\theta_0\|^2 \right] \,,
\end{align}
where $\Sigma = \E(x_i x_i^T) \in \reals^{d\times d}$ is the population covariance of the features distribution and $c_{\max}$ is the maximum singular value of $\Sigma$. 
We let $\cG$ be the probability event that \eqref{eq: estimation-error-bound} holds true. Then, by Proposition~\ref{estimation-error} we have $\prob(\cG) \ge 1-d^{-2} k^{-1.5} -2e^{-c_2 kd}$.

We are now in place to bound the regret of our policy.  Given that the length of episodes grow linearly, we have $\Regret(T) \le \sum_{k=1}^K \Reg_k$, with $\Reg_k$ denoting the total expected regret during episode $k$ and $K = \lfloor\sqrt{2T} \rfloor$. Consider the two cases below:
\begin{itemize}
\item $k\le c_0$: Here, $c_0$ is the constant in the statement of Proposition~\ref{estimation-error}. Since the benchmark prices are bounded by $P$, given by Lemma~\ref{boundP}, the regret at each step is at most $P$ and hence the total regret over such episodes is at most $P(\sum_{i=1}^{c_0} k + c_0d) < (c_0^2+c_0d)P $.
\item $k>c_0$: In this case, we write for $t\in E_k$,
\begin{align}
\E(\mreg_t) &\le c_3 c_{\max} C^2 N^5 \Big\{\E\Big[\Big(\|\hgamma^k-\gamma_0\|^2 + \| \htheta^k-\theta_0\|^2\Big) \cdot \mathbb{I}_\cG\Big] \nonumber\\
&\quad\quad\quad\quad\quad\quad\quad+ \E\Big[\Big(\|\hgamma^k-\gamma_0\|^2 + \| \htheta^k-\theta_0\|^2\Big) \cdot \mathbb{I}_{\cG^c}\Big] \Big\}\nonumber\\
&\le c_3 c_{\max} C^2 N^5 c_1 \frac{\log(kd)}{k} + 4W^2 \prob(\cG^c)\nonumber\\
&\le c_3 c_{\max} C^2 N^5 c_1 \frac{\log(kd)}{k} + 4W^2 (d^{-2} k^{-1.5} + 2e^{-c_2kd})\,.
\end{align}
Hence, the total expected regret over episode $k$ is bounded as follows
\begin{align}
\Reg_k = \sum_{t\in E_k} \E(\mreg_t) \le C \log(Td) (1+ d/k)\,,
\end{align}
where we used the fact that $|E_k| = k+d$, constant $C$ hides various constants.
\end{itemize}
To bound the cumulative expected regret up to time $T$, let $K = \lfloor \sqrt{2T}\rfloor$. By combining the two cases, we obtain
\[
\Regret(T) \le (c_0^2+c_0d)P + C  \log(Td) \sum_{k=1}^K (1+d/k ) \le \tilde{C}  \log(Td) ( \sqrt{T}+ d\log(T))\,.
\]  
This concludes the proof.
\subsection{Proof Theorem~\ref{thm:M3P-2}}\label{proof:thm-M3P-2}
Since we are treating $N$ (the maximum size of a consideration set) as constant and only interested in the lower bound of regret in terms of time horizon $T$, we consider the case of $N =1$. 
The $\Omega(\sqrt{T})$ follows by existence of the so-called `uninformative prices'~\cite{broder2012dynamic}. For a fixed time $t$, recall that $e^{u^0_t}/(1+e^{u^0_t})$ is the purchase probability, where $u^0_t = \<x_t,\theta_0\> - \<x_t,\gamma_0\> p_t$ and $x_t$ and $p_t$ are respectively the product feature and the posted price at time $t$. 
An uninformative price $p$, is any such price such that all the purchase probability curves (across model parameters) intersect at that price. The name comes from the fact that such price does not reveal any information about the underlying model parameters and hence does not help with the learning part (exploration of the space of the model parameters). Now, if an uninformative price is also the optimal price for a specific choice of parameters, then we would get a clear tension between the exploitation and exploration objectives. Indeed, for a policy to learn the underlying model parameters fast enough, it must necessarily choose prices that are away from the uninformative prices and this in turns leads to accruing regret when an uninformative price is in fact the optimal prices for the true model parameters. 

To construct uninformative prices, we let $\theta_{0,j} = 0$ and $\gamma_{0,j} = 0$ for $j>2$ and let $x_{t,1} = 1$ for all products. We then have 
\[
u^0_t = \<x_t,\theta_0\> -\<x_t,\gamma_0\> p_t = \theta_{0,1} - \gamma_{0,1} p_t\,.
\]
We set $\gamma_{0,1} = \theta_{0,1}$ and get $u^0_t = \theta_{0,1} (1-p_t)$. Therefore, $p_t = 1$, is an uninformative price, since the purchase probability curves only depend on $u^0_t$ (see Eq.~\eqref{probs}) and hence they all (across $\theta_{0,1}$) intersect at the common price $p_t =1$. 
In addition, for $\theta_{0,1} = 2$, it is easy to verify that $p_t = 1$ is the optimal price, using Proposition~\ref{propo:opt-price}.

Now that we have established the existence of uninformative prices, it can be shown that the worst-case $T$-regret of any policy is lower bounded by $\Omega(\sqrt{T})$.

In our next proposition, we make the above insight rigorous and formally states a lower bound on the regret of any policy. Before proceeding with the statement, we establish a lemma. 
We consider a \emph{problem class} $\mathcal{C}$ to be a pair $\mathcal{C} = (\mathcal{P}, {{\sf \Theta}})$, with ${{\sf \Theta}} = [\theta_{\min},\theta_{\max}]$ and $\mathcal{P} = [p_{\min}, p_{\max}]$, with $\theta_{\min}, p_{\min}\ge 0$. We would like to consider models with $\theta_{0,1} = \gamma_{0,1} = \theta\in {{\sf \Theta}}$ and pricing policies that set prices in $\mathcal{P}$. Note that in this case, a posted price $p$ yields the purchase probability $e^{\theta(1-p)/(1+e^{\theta(1-p)})}$. We also assume that for the optimal price under the model parameter $\theta$, denoted by $p^*(\theta)$, we have $p^*(\theta)\in\mathcal{P}$ for all $\theta\in {{\sf \Theta}}$. Our next lemma gives a sufficient condition for this assumption to hold.  

\begin{lemma}\label{lem:price-range}
Suppose that $p_{\min} \ge 1/\theta_{\max}$ and $p_{\max}\le \max(1, 2/\theta_{\min})$. Then, $p^*(\theta) \in \mathcal{P}$ for all $\theta\in {{\sf \Theta}}$.
\end{lemma}
We refer to Appendix~\ref{proof:lem-price-range} for the proof of Lemma~\ref{lem:price-range}.

\begin{propo}\label{pro:LB}
Define a problem class $\mathcal{C} = (\mathcal{P}, {{\sf \Theta}})$ by letting $\mathcal{P} = [2/5, 4/3]$ and ${{\sf \Theta}} = [3/2,5/2]$, and the purchase probability 
\[
q(p,\theta) = \frac{e^{\theta(1-p)}}{1+e^{\theta(1-p)}}\,.
\]
Then for any pricing policy $\pi$ setting prices in $\mathcal{P}$ and any $T\ge 2$, there exists a parameter $\theta\in {{\sf \Theta}}$ such that
\[
\Regret^\pi(\theta, T) \ge \frac{\sqrt{T}}{3 (41)^4}\,,
\]
where $\Regret^\pi(\theta, T)$ is the total expected regret of policy $\pi$ up to time $T$, under model parameter $\theta$.
\end{propo}
The proof of Proposition~\ref{pro:LB} follows is similar to the proof of \cite[Theorem 3.1]{broder2012dynamic}. However, that theorem applies only to the specific purchase probability function $1/2 + \theta - \theta p$. The proof of Proposition~\ref{pro:LB} requires some detailed analysis that is deferred to Appendix~\ref{proof:pro:LB}. 
\section{Proof of Proposition~\ref{propo:opt-price}}\label{proof:opt-price}

In the benchmark policy, the seller knows the model parameters $\theta_0, \gamma_0$. 
For simplicity, we use the shorthands $\beta_{i} = \<x_i,\gamma_0\>$, $e_{it}=\exp(\<x_i,\theta_0\> - \beta_i p_{it})$, and the sum as $G(e_t)= \sum_{\ell\in\cC_t}^{} e_{\ell t}$. The revenue function can be written in terms of $e_{it}$ as 
\[
\rev_t(p_t) = \sum_{i\in\cC_t}p_{it} \prob(i_t = i | \cC_t) = \sum_{i\in\cC_t} p_{it}\frac{e_{it}}{1+G(e_t)}\,,
\]
where we used~\eqref{probs}.
Writing the stationarity condition for the optimal price vector ${p}^*_t$, we get that for each $i\in\cC_t$:
\[
\frac{\partial \rev_t({p}^*_{t})}{\partial p_{it}} =\frac{e_{it}-e_{it}\beta_{i}p^*_{it}}{1+G({e}_{t})}+\frac{(\sum_{\ell\in\cC_t}p^*_{\ell t}e_{\ell t})e_{it}\beta_{i}}{(1+G({e}_{t}))^{2}} = 0\,,
\]
which is equivalent to
\[
\beta_{i}\frac{e_{it}}{1+G({e}_t)}\bigg\{\frac{1}{\beta_{i}}-p^*_{it}+\underbrace{\frac{\sum_{\ell\in\cC_t}p^*_{\ell t}e_{\ell t}}{1+G({e}_t)}}_{\rev_t(p_t)} \bigg\}= 0\,.
\]
Since $e_{it}>0$, the above equation implies that
\begin{align}\label{opt-price-z}
p^*_{it}=\frac{1}{\beta_i}+{\rev_t({p}^*_t)}\,.
\end{align}
Define $B^0_t \equiv \rev_t({p}^*_t)$. We next show that $B^0_t$ is the solution to Equation~\eqref{eq:opt-price}.
By multiplying both sides of~\eqref{opt-price-z} by $e_{\ell t}$ and summing over $\ell\in \cC_t$, we have 
\begin{align*}
\sum_{\ell\in\cC_t}e_{\ell t}p^*_{\ell t} &= \sum_{\ell\in\cC_t}\frac{e_{\ell t}}{\beta_{\ell}}+ B^0_t \Big(\sum_{\ell\in\cC_t} e_{\ell t}\Big)
= \sum_{\ell\in\cC_t}\frac{e_{\ell t}}{\beta_{\ell}} + B^0_t G({e}_t)\,. 
\end{align*}
By definition of $B^0_t$, the left-hand side of the above equation is equal to $B^0_t(1+G({e}_t))$. By rearranging the terms we obtain
\begin{align*}
B^0_t &=\sum_{\ell\in\cC_t}\frac{e_{\ell t}}{\beta_{\ell }}=\sum_{\ell\in\cC_t}\frac{1}{\beta_{\ell}}e^{\<x_{\ell},\theta_0\>-\beta_{\ell}p_{\ell t}}\\
&=\sum_{\ell\in\cC_t}\frac{1}{\beta_{\ell}}\exp\Big\{{\<x_{\ell},\theta_0\>-\beta_{\ell}\Big(\dfrac{1}{\beta_{\ell}}+B^0_t\Big)}\Big\}\\
&=\sum_{\ell\in\cC_t}\frac{1}{\beta_{\ell}}e^{\<x_{\ell},\theta_0\>}e^{-(1+\beta_{\ell}B^0_t)}\,,
\end{align*}
where the second line follows from Equation~\eqref{opt-price-z}.

Regarding the uniqueness of the solution of~\eqref{eq:opt-price}, note that the left-hand side of~\eqref{eq:opt-price} is strictly increasing in $B$ and is zero at $B=0$, while the right hand side is strictly decreasing in $B$ and is positive at $B=0$. Therefore, Equation~\eqref{eq:opt-price} has a unique solution.
\section{Proof of Lemma~\ref{lemma:price-lipschitz}}
Define function $f:\reals\times \reals^N \times \reals^N \mapsto \reals$ as
\begin{align}\label{f:def}
f(B,{\delta},{\beta}) \equiv B - \sum_{\ell\in\cC_t} \frac{1}{\beta_\ell} e^{\delta_\ell} e^{-(1+\beta_\ell B)}\,.
\end{align}
By characterization of the pricing function $g$, given in Proposition~\ref{propo:opt-price}, we have $p^{(1)}_{it} = \frac{1}{\beta^{(1)}_i} + B^{(1)}_t$ and $p^{(2)}_{it} = \frac{1}{\beta_i^{(2)}} + B^{(2)}_t$, where $B^{(1)}_t$ and $B^{(2)}_t$ are the solution of $f(B,{X}_t {\theta}_1, X_t \gamma_1) = 0$ and $f(B,{X}_t {\theta_2},X_t \gamma_2) = 0$.
 
 By implicit function theorem for a point $(B,{\delta},{\beta})$ that satisfies $f(B,{\delta}, {\beta})$ = 0, there exists an open set around $({\delta}, {\beta})$, and a unique differentiable function $h:U\mapsto \reals$ such that $h({\delta},{\beta}) = B$ and $f(h({z}_1,{z_2}), {z}_1, {z_2}) = 0$ for all $({z}_1, {z}_2)\in U$.  Furthermore, the partial derivative of $g$ can be computed as
 \begin{align*}
\frac{\partial h}{\partial \delta_{i}} ({\delta},{\beta})&=-\Big[\frac{\partial f}{\partial B}({\delta,\beta})\Big]^{-1}\frac{\partial f}{\partial \delta_i} (h({\delta},{\beta}), {\delta},{\beta})\\
& =-\Big(1+\sum_{\ell\in\cC_t}e^{\delta_{\ell}}e^{-(1+\beta_{\ell}B)}\Big)^{-1} \Big(-\frac{1}{\beta_i}e^{-(1+\beta_i B)}e^{\delta_{i}}\Big )\\
& < \frac{e^{\delta_i}}{\beta_i}<\frac{e^W}{L_0}\,,
\end{align*}
 where in the last step we use the normalization $|\delta_i| = |\<{x}_i,{\theta}_1\>|\le \l1u$, and $0< L_0 < \min \beta_i$ is the lower bound on the price sensitivities. Likewise, we have
  \begin{align*}
\frac{\partial h}{\partial \beta_{i}} ({\delta},{\beta})&=-\Big[\frac{\partial f}{\partial B}({\delta,\beta})\Big]^{-1}\frac{\partial f}{\partial \beta_i} (h({\delta},{\beta}), {\delta},{\beta})\\
& =-\Big(1+\sum_{\ell\in\cC_t}e^{\delta_{\ell}}e^{-(1+\beta_{\ell}B)}\Big)^{-1} \Big(\Big(\frac{B}{\beta_i}+ \frac{1}{{\beta_i}^2}\Big)e^{-(1+\beta_i B)}e^{\delta_{i}}\Big )\\
& <\Big(\frac{B}{L_0}+ \frac{1}{{L_0}^2}\Big) e^{\delta_i}<(N e^{W-1}+1) \frac{e^W}{L_0^2}\,,
\end{align*}
 where we used the fact that the solution $B$ of $f(B,\delta, \beta)$ satisfies $B\le N e^{W-1}/L_0$. (This follows readily by noting that the right-hand side of \eqref{f:def} is non-increasing in $B$.)  
 This shows that $g({\delta}, {\beta})$ is a Lipschitz function of ${\delta}$, with Lipschitz constant $CN$, where $C\equiv  N e^{2W}/\min(L_0^2,1)$.
 Therefore,
 \begin{align}
 |p^{(1)}_{it} - p^{(2)}_{it}| &= {|B^{(1)}_t - B^{(2)}_t|} + \Big|\frac{1}{\beta_i^{(1)}} - \frac{1}{\beta_i^{(2)}} \Big|\nonumber\\
  & = |h(X_t{\theta}_1, X_t \gamma_1) - h(X_t{\theta}_2, X_t \gamma_2)| + \frac{1}{\beta_i^{(1)}\beta_i^{(2)}} |\<x_i, \gamma_1 -\gamma_2\>|\nonumber\\
 &\le C N \left(\|{X}_t({\theta_1} - {\theta_2})\|_1 +  \|{X}_t(\gamma_1-\gamma_2)\|_1\right) + \frac{1}{L_0^2} |\<x_i, \gamma_1 -\gamma_2\>| \,.
 \end{align}
 By Cauchy-Schwarz inequality we have $\|{X}_t({\theta_1} - {\theta_2})\|_1 \le \sqrt{N} \|{X}_t({\theta_1} - {\theta_2})\|$ and $\|{X}_t({\gamma_1} - {\gamma_2})\|_1 \le \sqrt{N} \|{X}_t({\gamma_1} - {\gamma_2})\|$. 
 Hence, $\|p^{(1)}_t - p^{(2)}_t\| \le   \tilde{C} N^2 \left( \|X_t(\theta_1-\theta_2)\| + \|X_t(\gamma_1-\gamma_2)\|\right) $, for some constant $\tilde{C}$.
 The claim now follows by using $a+b < \sqrt{2(a^2+b^2)}$.

\section{Proof of Proposition~\ref{estimation-error}}

We start by recalling the notation $\nu_0 = (\theta_0^T, \gamma_0^T)^T$ and define
$\tilde{X}_t = [{X}_t,\,  -\diag(p_t) X_t]$.
To prove Proposition~\ref{estimation-error}, we first rewrite the loss function in terms of the augmented parameter vector $\nu$. (Recall our convention that  $\emptyset$ corresponds to ``no-purchase" with $u^0_{\emptyset t}(\cdot)=0$ .)

\begin{align}
\begin{split}
\mathcal{L}_k({\nu})&=-\frac{1}{kd}{\displaystyle \sum_{t\in I_{k}}\log\frac{\exp(u^0_{i_tt})}{\sum_{\ell\in\cC_t\cup\{\emptyset\}}\exp(u^0_{\ell t})}}
\\
&=\frac{1}{kd}\sum_{t\in I_{k}}(\log(1+\sum_{\ell\in\cC_t}e^{\<{x}_{\ell},{\theta}\>-p_{\ell t}\<{x}_{\ell},{\gamma}\>})-\left(\begin{cases}
0 & i_{t}=\emptyset\\
\<x_{i_t},{\theta}\>-p_{i_t}\<{x}_{i_t},{\gamma}\>& \text{otherwise}
\end{cases}\right)\\
&=\frac{1}{kd}\sum_{t\in I_k}\left\{\log(1+\sum_{\ell\in\cC_t}e^{\<\tilde{x}_{\ell},{\nu}\>})-\left(\begin{cases}
0 & i_{t}=\emptyset\\
\<\tilde{x}_{i_t},{\nu}\> & \text{otherwise}
\end{cases}\right)\right\}\,,
\end{split}
\end{align}
where $\tilde{x}_{\ell}=[{x}_{\ell}^T,
-p_{\ell t}{x}_{\ell}^T]^T$.
The gradient and the hessian of $\mathcal{L}_k$ are given by 
\begin{align}\label{grad0}
\nabla\mathcal{L}_k({\nu})=\frac{1}{kd}\sum_{t\in I_k}\left(\frac{\sum_{\ell\in\cC_t} \exp(u^0_{\ell})u^0_{\ell}\tilde{x}_{\ell}}{1+\sum_{\ell\in\cC_t}\exp(u^0_{\ell})}-\tilde{x}_{i_{t}}\right)\,,
\end{align}

\begin{align}
\nabla^{2}\mathcal{L}_k({\nu})=\frac{1}{kd}\sum_{t\in I_k}\frac{(1+\sum_{\ell\in\cC_t}\exp(u^0_{\ell}))(\sum_{\ell\in\cC_t}\exp(u^0_{\ell})((u^0_{\ell})^{2}+1)\tilde{x}_{\ell}^{\otimes 2})-(\sum_{\ell\in\cC_t}u^0_{\ell}\tilde{x}_{\ell})^{\otimes2}}{(1+\sum_{\ell\in\cC_t}\exp(u^0_{\ell}))^{2}}\,.
\end{align}

We proceed by bounding the gradient and the hessian of the loss function. Before that, we establish an upper bound on the prices that are set by the pricing function $g$.
\begin{lemma}\label{boundP}
Suppose that $\|{x}_\ell\|_\infty \le 1$ and {{$\|\nu_0\|_1\le \l1u$}}. Let $B^u = B^u(W, L_0, N)$ be the solution to the following equation:
\begin{align}\label{eq:opt-price-last}
B = N \frac{1}{L_0} e^{-(1+L_0 B)} e^{\l1u}\,. 
\end{align}
Then, the prices set by the pricing function $p_t = g(X_t \gamma, {X} {\theta})$, where {{$\nu_0 = (\theta_0^T,\gamma_0^T)^T$}}, are bounded by $P=1/L_0 + B^u$. 
\end{lemma}
The proof of above Lemma follows readily by noting that the right-hand side of \eqref{eq:opt-price-last} is an upper bound for the right hand side of \eqref{eq:opt-price} and therefore $B^0_t\le B^u$. The results then follows by recalling that the pricing function sets prices as $p_{it}= 1/\beta_i + B^0_t$.

To bound the gradient of the loss function at the true model parameters, note that
\begin{eqnarray}\label{mygrad}
 \nabla\mathcal{L}_k(\nu_0)  = \frac{1}{kd} \sum_{t\in I_k}  S_t\,, \quad \text{ with }\quad  S_t \equiv \frac{\sum_{\ell\in\cC_t}\exp(u^0_{\ell t})u^0_{\ell t}\tilde{x}_\ell}{1+\sum_{\ell\in\cC_t}\exp(u^0_{\ell t})}-\tilde{x}_{i_{t}}
\end{eqnarray}
We also have
\begin{align}\label{eq:u}
|u_{\ell t}^0|\leq|\<x_{\ell},\theta_0\> + |\<{x}_\ell,\gamma_0\>| p_{\ell t} \leq W(1+P)   \equiv {M}\,,
\end{align}
for a constant $M = M(W,L_0, N)>0$ and so $\|S_t\|\le (M+1)\sqrt{d(1+P^2)}$, because $\|\tilde{x}_\ell\|\le \sqrt{d(1+P^2)}$. Note that by~\eqref{mygrad}, $\nabla \mathcal{L}_k(\nu_0)$ is written as some of $kd$ terms. In each term, the index $i_t$ has randomness coming from the market noise distribution. By a straightforward calculation, one can verify that each of these terms has zero expectation. Using~\eqref{eq:u} and by applying Matrix Freedman inequality to the right-hand side of~\eqref{mygrad}, followed by union bounding over $d$ coordinates of feature vectors, we obtain
\begin{align}\label{grad2}
    \left\Vert \nabla\mathcal{L}_k(\nu_0)\right\Vert \leq \lambda_k,\quad \text{with } \lambda_k \equiv 2(M+1)\sqrt{ d(1+P^2) \frac{\log(d|I_k|)}{|I_k|}}\,,
\end{align}
with probability at least $1-d^{-0.5} |I_k|^{-1.5}$. 

We next pass to lower bonding the hessian of the loss. For any $\tilde{\nu}$ with $\|\tilde{\nu}\|\le W$, we have
\begin{align}
&\<\nu_0-\hpsi^k,\nabla^2 \mathcal{L}_k(\tilde{\nu})(\nu_0-\hpsi^k)\>\nonumber\\
    & = \frac{1}{kd}\sum_{t\in I_k}
   \frac{(1+\sum_{\ell \in\cC_t}\exp(\tilde{u}^0_{\ell t}))(\sum_{\ell\in\cC_t}\exp(\tilde{u}^0_{\ell t})((\tilde{u}^0_{\ell t})^{2}+1)(\tilde{x}_{\ell}(\nu_0-\hpsi^k))^{2}-(\sum_{\ell \in\cC_t}\tilde{u}^0_{\ell t}\tilde{x}_{\ell}(\nu_0-\hpsi^k))^{2}}{(1+\sum_{\ell \in\cC_t}\exp(\tilde{u}^0_{\ell t}))^{2}}\nonumber\\
   & \stackrel{(a)}{\geq} \frac{1}{kd}\sum_{t\in I_k}\frac{\sum_{\ell\in\cC_t}\exp(\tilde{u}^0_{\ell t})((\tilde{u}^0_{\ell t})^{2}+1)(\tilde{x}_{\ell}(\nu_0-\hpsi^k))^{2}+\sum_{\ell\in\cC_t}\exp(\tilde{u}^0_{\ell t})\sum_{\ell\in\cC_t}\exp(\tilde{u}^0_{\ell t})(\tilde{x}_{\ell}(\nu_0-\hpsi^k))^{2}}{(1+\sum_{\ell\in\cC_t}\exp(\tilde{u}^0_{\ell t}))^{2}}\nonumber\\
    & \stackrel{(b)}{>} \frac{1}{kd}\sum_{t\in I_k}\Big(\frac{\sum_{\ell\in\cC_t}\exp(\tilde{u}^0_{\ell t})(\tilde{x}_{\ell}(\nu_0-\hpsi^k))^{2}}{1+\sum_{\ell\in\cC_t}\exp(\tilde{u}^0_{\ell t})}\Big)\nonumber\\
  & \stackrel{(c)}{\geq} \frac{e^{-{M}}}{kd}\sum_{t\in I_k}\frac{\sum_{\ell\in\cC_t}(\tilde{x}_{\ell}(\nu_0-\hpsi^k))^{2}}{1+Ne^{{M}}}\nonumber\\
   & \stackrel{(d)}{\geq} \frac{e^{-{M}}}{kd(1+N e^{{M}})} \Vert \boldsymbol{\tilde{X}}^{(k-1)} (\gamma_0-\hgamma^k) \Vert ^{2}\nonumber\\
 &\equiv \frac{c_0(W,L_0,N)}{N kd} \Vert \boldsymbol{\tilde{X}}^{(k-1)} (\nu_0-\hpsi^k) \Vert ^{2}\,,
 \label{eq:bound-D}
\end{align}
where (a) and (b) follow from Jensen's Inequality. (c) is because $|\tilde{u}_{\ell}^0|\leq {M}$ by definition~\eqref{eq:u} and the assumption $\|\tilde{\nu}\|\le W$; in $(d)$,  we define $n_k \equiv \sum_{t\in I_k} |\cC_t| \le N kd$ and construct $\boldsymbol{\tilde{X}}^{(k-1)} $ of size $n_k$ by $2d$,  by staking the features $\tilde{x}_t$, for $t\in I_k$, row-wise. 

 By optimality of $\hpsi^k$ and the second order Taylor expansion we have
\begin{align}\label{oracle}
0\ge \cL(\nu_0) - \cL(\hpsi^k) = -\<\nabla \cL(\nu_0), \hpsi^k - \nu_0\> - \frac{1}{2} \<\hpsi^k-\nu_0,\nabla^2 \cL(\tilde{\nu}) (\hpsi^k - \nu_0))\>\,,
\end{align}
for some $\tilde{\nu}$ on the segment between $\nu_0$ and $\hpsi^k$.

Therefore by~\eqref{oracle} and \eqref{eq:bound-D} we arrive at
\[
\frac{c_0(W,L_0,N)}{2Nkd} \|\boldsymbol{\tilde{X}}^{(k-1)}(\nu_{0}- \hpsi^k)\|^2  \le \|\nabla \cL(\nu_0)\| \|\hpsi^k - \nu_0\|.
\] 
Using the bounds on the gradient  given by~\eqref{grad2}, we get that with probability at least $1 - d^{-0.5}|I_k|^{-1.5}$
\begin{align}\label{eq:tX-oracle1}
\frac{c_0(W,L_0,N)}{2N kd} \|\boldsymbol{\tilde{X}}^{(k-1)}(\nu_0-\hpsi^k)\|^2 \le \lambda_k \|\hpsi^k - \nu_0\| \,.
\end{align}
Next, in order to lower bound the left-hand side of \eqref{eq:tX-oracle1}, we first lower bound the minimum eigenvalue of $\hSigma_k$, the empirical second moment of $\boldsymbol{\tilde{X}}^{(k-1)}$, defined as
\[
\hSigma_k \equiv\frac{1}{n_k} \boldsymbol{\tilde{X}}^{(k-1)} (\boldsymbol{\tilde{X}}^{(k-1)})^T  = \frac{1}{n_k} \sum_{t\in I_k, \ell\in \cC_t} \tilde{x}_\ell \tilde{x}_\ell^T\,,
\]
where we recall that $\tilde{x}_\ell = [x_\ell^T, -p_{\ell t} x_\ell^T]^T$ for $\ell\in \cC_t$.

 Since rows of $\boldsymbol{\tilde{X}}^{(k-1)}$ are bounded, they are subgaussian. Using~\cite[Remark 5.40]{vershynin2010introduction}, there exist universal constants $c$ and $C$ such that for every {$m\ge0$}, the following holds with probability at least $1 - 2e^{-c{m}^2}$:
\begin{align}
\Big\|\hSigma_k - \tilde{\Sigma} \Big\|_{\rm op} ~\le~ \max(\delta,\delta^2) \quad \text{where} \quad \delta = C\sqrt{\frac{d}{n_k}} + \frac{{m}}{\sqrt{n_k}}\,, 
\end{align}
where {$\tilde{\Sigma} = \E[\tilde{x}_\ell \tilde{x}_\ell^\sT] \in \reals^{2d\times 2d}$} is the covariance of the these vectors. 
We proceed by computing $\tilde{\Sigma} \equiv \E[\tilde{x}_\ell \tilde{x}_\ell^T]$. Recall that for period $t\in I_k$, the prices are set uniformly at random from $[0,B]$. Hence, letting $\mu = \E[x_\ell]$  and $\Sigma \equiv \E[x_\ell x_\ell^T]$ the population first and second moments of $x_\ell$, we have
\begin{align}\label{schur}
\tilde{\Sigma} \equiv \E[\tilde{x}_\ell \tilde{x}_\ell^T] = \begin{pmatrix}
\Sigma &\frac{-B\mu}{2}\\
-\frac{B\mu^T}{2} & \frac{B^2}{3}
\end{pmatrix}\,.
\end{align} 
The Schur complement of $\tilde{\Sigma}$, with respect to block $\Sigma$, reads as
\[
\frac{B^2}{3} - \frac{B^2}{4} \mu^T \Sigma^{-1}\mu \ge \frac{B^2}{3} - \frac{B^2}{4} = \frac{B^2}{12}\,,
\] 
where we used the fact that $\mu^T \Sigma^{-1}\mu\le 1$ which follows readily by looking at the Schur complement of the second moment of $a = [x_t;1]$, i.e., $\E[aa^T] = [\Sigma \;\;\mu; \mu^T\;\; 1]$ and use the fact that the Schur complement of positive semidefinite matrices is nonnegative. As a result of~\eqref{schur}, the singular values of $\tilde{\Sigma}$ are larger than $\tilde{c}_{\min} \equiv \min(c_{\min}, B^2/12)$.

Then, for $n_k > c_0 d$, with probability at least $1 - 2e^{-c_2 n_k}$, the following is true:
\begin{align}\label{eventG}
\Big\|\hSigma_k- \tilde{\Sigma} \Big\|_{\rm op} \le \frac{1}{2} \tilde{c}_{\min}\,,
\end{align}
and hence the minimum singular value of $\hSigma_k$ is at least $\tilde{c}_{\min}/2$. Using this in Equation~\eqref{eq:tX-oracle1} we get that with probability at least $1 - d^{-0.5} |I_k|^{-1.5} - 2e^{-c_2 n_k}$,
\begin{align}\label{eq:prediction}
\frac{1}{2} \tilde{c}_{\min} \|\hpsi^{k} - \nu_0\|^2 ~\le~ \frac{1}{n_k}  \|\boldsymbol{\tilde{X}}^{(k-1)}(\hpsi^{k} - \nu_0)\|^2\,.
\end{align}
Combining~\eqref{eq:prediction} and \eqref{eq:tX-oracle1} leads to
\begin{align}\label{eq:prediction}
\frac{n_kc_0(W,L_0,N)}{4Nkd} \tilde{c}_{\min} \|\hpsi^{k} - \nu_0\|^2 ~&\le~ \frac{c_0(W,L_0,N)}{2Nkd}  \|\boldsymbol{\tilde{X}}^{(k-1)}(\hpsi^{k} - \nu_0)\|^2
 \le  \lambda_k\|\nu_0-\hpsi^k\|\,,
\end{align}
which gives that with probability at least $1 - d^{-2} k^{-1.5} - 2e^{-c_2 kd}$, the following holds true
\[
\|\hpsi^k - \nu_0\| \le \frac{4Nkd}{n_k c_0 \tilde{c}_{\min}} \lambda_k \le \frac{4N}{c_0 \tilde{c}_{\min}} \lambda_k\,.
\]
The proof is complete.
\section{Proof of Lemma~\ref{lemma:regret-lipschitz}}
In this lemma, we aim at bounding the revenue loss (against the clairvoyant policy) in terms of the distance between the posted price vector and the optimal one posted by the clairvoyant policy. 
By Taylor expansion, 
\begin{align}\label{eq:taylor}
\rev_t(p_t)= \rev_t(p_t^{*}) + \nabla \rev_t({p}^{*}_t)({p}_t-{p}^{*}_t)+\frac{1}{2}({p}_t-{p}^{*}_t)^{T}\nabla^{2}\rev_t(\tilde{{p}})({p}_t-{p}^{*}_t)\,,
\end{align}
for some $\tilde{p}$ between $p_t$ and $p^{*}_t$. Note that $p^{*}_t = \arg \max \rev_t(p)$, thus $\nabla \rev_t({p}^{*}_t)=0$ and the first term in the Taylor expansion vanishes.

In order to prove the result, it suffices to show that the operator norm $\|\nabla^{2}\rev_t(\tilde{{p}})\|_2$ is bounded. 
Fix $i,j \in \cC_t$. We have
\begin{align}
\frac{\partial \rev_t(p)}{\partial p_{i}} = \frac{e^{u^0_{it}}(1- \beta_i p_{i})}{1+\sum_{\ell\in\cC_t} e^{u^0_{\ell t}}} +  \sum_{k\in \cC_t} \frac{p_{k}e^{u^0_{kt}}}{(1+\sum_{\ell\in\cC_t} e^{u^0_{\ell t}})^2}\,,
\end{align}
with $\beta_i = \<x_i,\gamma_0\>$. Taking derivative with respect to $p_{j}$, we get
\begin{align}
{\frac{\partial^2 \rev_t(p)}{\partial p_{i} \partial p_{j}} =  \frac{e^{u^0_{jt}}}{(1+\sum_{\ell\in\cC_t} e^{u^0_{\ell t}})^2} \Big[ \beta_j(1-\beta_i p_{i}) e^{u^0_{it}} + 1 - p_{j} \beta_j \Big]
 + 2\beta_j e^{u^0_{jt}} \frac{\sum_{k\in\cC_t}p_{k} e^{u^0_{kt}}}{ (1+\sum_{\ell\in\cC_t} e^{u^0_{\ell t}})^3}}\,. 
\end{align}
By Lemma~\ref{boundP}, we have $p_{it} \le P$. Also, by~\eqref{eq:u}, we have $|u^0_{\ell t}|\le M$. In addition, $0\le \beta_i \le \|x_i\|_\infty \|\gamma_0\|_1 \le W$. Since $P$ and $M$ are constants depending only on $W$ and $L_0$, there exists a constant $c_1(W,L_0) >0$, such that 
\begin{align}\label{hessian-infty}
\Big|\frac{\partial^2 \rev_t}{\partial p_{i} \partial p_{j}} \Big| \le c_1(W,L_0)\,,
\end{align}
uniformly over $i,j \in \cC_t$. We next bound the operator norm of $\nabla^2 \rev_t(p)$. Note that for a matrix $A\in \reals^{N\times N}$, we have
\begin{align}
\|A\|_{2} = \sup_{\| {u}\| \leq 1} {|{u}^T A {u}|}
\le \sup_{\| {u}\| \leq1} \Big\{\sum_{i,j=1}^N |A_{i,j}|\,  |u_i|\,  |u_j| \Big\}
\le |A|_\infty \sup_{\| {u}\| \leq1} \|{u}\|_1^2 
\leq N |A|_{\infty}\,,\label{eq:1}
\end{align}
where $|A|_\infty = \max_{1\le i,j\le N} |A_{ij}|$. Therefore, the result follows by using~\eqref{hessian-infty}.

\section{Proof of Proposition~\ref{pro:LB}}\label{proof:pro:LB}
%
Define the purchase probability $q(\theta,p)$ and the revenue function $r(\theta,p)$ as 
\begin{align}
q(p;\theta) = \frac{e^{\theta(1-p)}}{1+e^{\theta(1-p)}}, \quad  r(p;\theta) = p \frac{e^{\theta(1-p)}}{1+e^{\theta(1-p)}}\,.
\end{align}
We next find the optimal price corresponding to parameter $\theta$. Write
\begin{align}
r'(p;\theta) = 1 - (1+\theta p) \frac{1}{1+e^{\theta(1-p)}} + \frac{\theta p}{(1+e^{\theta(1-p)})^2}\,,
\end{align}
which implies that the optimal price $p^*(\theta)$ satisfies the following relation
\begin{align}\label{optimal-price}
\theta p^*(\theta) = 1+ e^{\theta(1-p^*(\theta))}\,.
\end{align}



Our next lemma is on some properties of the problem class $\mathcal{C}$ considered in the statement of Proposition~\ref{pro:LB}. Its proof is given in Appendix~\ref{proof:lem:properties}.
\begin{lemma}\label{lem:properties}
For all $p\in \mathcal{P} = [2/5, 4/3]$ and $\theta\in {{\sf \Theta}} = [3/2,5/2]$,
\begin{enumerate}
\item[(i)] For $\theta_0 = 2$, we have $|q(p,\theta) - q(p,\theta_0)| \le |p^*(\theta) - p|\, |\theta  - \theta_0|$.  
\item[(ii)] $r(p^*(\theta)) - r(p) \ge \frac{1}{520}(p^*(\theta) - p)^2$.
\item[(iii)] $|p^*(\theta) - p^*(\theta_0)| \ge 0.2 |\theta-\theta_0|$.
\end{enumerate}
\end{lemma}

In order to establish a lower bound on the regret of a policy, we need to quantitatively measure the uncertainty of the policy about the unknown model parameter $\theta$. 
To this end, we leverage the notion of the KL divergence of two probability measures $f_0$ and $f_1$ defined on a discrete sample space $\Omega$ as
\begin{align}
\KL(f_0;f_1) \equiv \sum_{\omega\in \Omega} f_0(\omega) \log\left(\frac{f_0(\omega)}{f_1(\omega)}\right)\,.
\end{align}
For any pricing policy $\pi$ and parameter $\theta\in {{\sf \Theta}}$ we let $f_t^{\pi,\theta}: \{0,1\}^t \to [0,1]$ be the probability distribution of the customer purchase responses $\bY_t = (Y_1, \dotsc, Y_t)$ under pricing policy $\pi$ and model parameter $\theta$. Formally, for all $\by_t = (y_1, \dotsc, y_t)\in \{0,1\}^t$,
\[
f_t^{\pi,\theta}(\by_t) = \prod_{i=1}^t q(p_i,\theta)^{y_i} (1- q(p_i,\theta))^{1-{y_i}}\,,
\] 
where $p_i = \pi(\by_{i-1})$ is the price posted under policy $\pi$.

The next lemma shows that in order to reduce the uncertainty about model parameter $\theta_0$ (equivalently increasing $\KL(f_t^{\pi,\theta_0};f_t^{\pi,\theta})$), the policy incurs a large regret. We refer to Appendix~\ref{proof:lem-uncertainty} for its proof.
\begin{lemma}\label{lem:uncertainty}
For any $\theta\in {{\sf \Theta}}$, $t\ge 1$, $\theta_0 = 2$ and any policy $\pi$ setting prices in $\mathcal{P}$, we have 
\[
\KL(f_t^{\pi,\theta_0};f_t^{\pi,\theta}) \le {3640}(\theta_0-\theta)^2  \Regret^\pi(t,\theta_0)\,,
\]
where $\Regret^\pi(t,\theta_0)$ indicates the total expected regret of the pricing policy $\pi$, up until step $t$, under model parameter $\theta_0$.
\end{lemma}

The next lemma is similar to~\cite[Lemma 3.4]{broder2012dynamic} and its proof follows along the same lines for our purchase probability function (in particular, using Lemma~\ref{lem:properties} ($ii$, $iii$)). 
\begin{lemma}\label{lem:cost}
Let $\pi$ be any pricing policy setting prices in $\mathcal{P}$. Then, for any $T\ge 2$ and model parameters $\theta_0 = 2$ and $\theta_1 = \theta_0 + \frac{1}{4}T^{-1/4}$, we have
\[
\Regret^\pi(T,\theta_0) + \Regret^\pi(T,\theta_1) \ge \frac{\sqrt{T}}{2080(41)^2} e^{-\KL(f_T^{\pi,\theta_0}; f_T^{\pi,\theta_1})}\,.
\]
\end{lemma}

Armed with Lemma~\ref{lem:uncertainty} and \ref{lem:cost} we are ready to prove Theorem~\ref{thm:M3P-2}. Let $\theta_0 = 2$ and $\theta_1 = \theta_0 + \frac{1}{4}T^{-1/4}$. Then, by non-negativity of KL-divergence and by virtue of Lemma~\ref{lem:uncertainty} we have
\[
\Regret^\pi(T,\theta_0) + \Regret^\pi(T,\theta_1) \ge \frac{16}{3640} \sqrt{T}\; \KL(f_t^{\pi,\theta_0};f_t^{\pi,\theta})\,.
\]
Combining this bound with Lemma~\ref{lem:cost} we get
\begin{align*}
2\left\{ \Regret^\pi(T,\theta_0) + \Regret^\pi(T,\theta_1)\right\} &\ge \frac{16}{3640} \sqrt{T}\; \KL(f_t^{\pi,\theta_0};f_t^{\pi,\theta}) +  \frac{\sqrt{T}}{2080(41)^2} e^{-\KL(f_T^{\pi,\theta_0}; f_T^{\pi,\theta_1})}\\
&\ge \frac{\sqrt{T}}{2080(41)^2} \left\{\KL(f_t^{\pi,\theta_0};f_t^{\pi,\theta}) + e^{-\KL(f_t^{\pi,\theta_0};f_t^{\pi,\theta}) } \right\} \ge \frac{\sqrt{T}}{2080(41)^2}\,,
\end{align*}
where we used that $a+e^{-a}\ge 1$ for all $a\ge 0$. Hence,
\begin{align*}
\max_{\theta\in \{\theta_0,\theta_1\}} \Regret^\pi(T,\theta)  \ge \frac{\sqrt{T}}{2\times 2080(41)^2} >\frac{\sqrt{T}}{3(41)^4} \,,
\end{align*}
which completes the proof.

\subsection{Proof of Lemma~\ref{lem:price-range}}\label{proof:lem-price-range}
By~\eqref{optimal-price} we have $\theta p^*(\theta) \ge 1$. Hence, for $\theta\in {\sf \Theta}$ we have $p^*(\theta)\ge 1/\theta_{\max}$. To prove the other side, assume that $p^*(\theta) >1$ for some $\theta \in {{\sf \Theta}}$. Then, by \eqref{optimal-price}, we have
\[
\theta p^*(\theta) = 1+ e^{\theta(1-p^*(\theta))} < 2\,,
\]
which gives $p^*(\theta) < 2/\theta \le 2/\theta_{\min}$. As a result, $p^*(\theta)\le \max(1, 2/\theta_{\min})$ completing the proof of the lemma.
\subsection{Proof of Lemma~\ref{lem:properties}}\label{proof:lem:properties}
To prove the first item, write
\begin{align*}
|q(p,\theta) - q(p,\theta_0)| &= \Big|\frac{1}{1+e^{\theta(1-p)}} - \frac{1}{1+e^{\theta_0(1-p)}} \Big|\\
&\le \frac{|e^{\theta(1-p)} - e^{\theta_0(1-p)}|}{(1+e^{\theta(1-p)}) (1+e^{\theta_0(1-p)})}\\
&\le \frac{e^{\max(\theta,\theta_0)|1-p|} (1 -  e^{-|\theta_0-\theta| |1-p|})}{(1+e^{\theta(1-p)}) (1+e^{\theta_0(1-p)})}\\
&\le 1 -  e^{-|\theta_0-\theta| |1-p|} \le |\theta_0-\theta|\, |1-p|\\
&= |\theta_0-\theta|\, |p^*(\theta_0)-p|\,,
\end{align*}
where the last inequality follows since $1-e^{-x} \le x$ for $x\ge 0$. Further, the last inequality holds since $p^*(\theta_0) = 1$ for $\theta_0 = 2$ by using~\eqref{optimal-price}.

To prove the second item, by some algebraic calculation, 
\begin{align}\label{2derivative}
r''(p) = \theta e^{\theta(1-p)} \left(\theta p (1-e^{\theta(1-p)}) -2 (1+e^{\theta(1-p)}) \right) (1+e^{\theta(1-p)})^{-3}\,.
\end{align}
Since $\theta\in {{\sf \Theta}}$ and $p\in \mathcal{P}$, if $p\le 1$ then $\theta p (1-e^{\theta(1-p)}) -2 (1+e^{\theta(1-p)}) \le -4$; otherwise
\begin{align*}
\theta p (1-e^{\theta(1-p)}) -2 (1+e^{\theta(1-p)}) &\le \frac{10}{3}(1-e^{\theta(1-p)}) -2 (1+e^{\theta(1-p)})\\
&\le \frac{4}{3} - \frac{16}{3} e^{\theta(1-p)} \le \frac{4}{3} - \frac{16}{3} e^{-5/6} < -0.98\,.
\end{align*}
Combining these two cases and using the characterization~\eqref{2derivative}, we arrive at
\begin{align}\label{2-der}
r''(p) \le -0.98 \theta e^{\theta(1-p)} (1+e^{\theta(1-p)})^{-3} \le -0.98\times 3/2\times e^{-5/6} \times (1+e^{3/2})^{-3} < -1/260\,,
\end{align} 
where we used that $e^{\theta(1-p)}\in [e^{-5/6}, e^{3/2}]$ for $\theta\in {{\sf \Theta}}$ and $p\in \mathcal{P}$.

Now by Taylor expansion of $r(p)$ around $p^*$ we obtain
\[
r(p) = r(p^*(\theta)) + r'(p^*(\theta)) (p - p^*(\theta)) + \frac{1}{2} r''(\tilde{p}) (p-p^*(\theta))^2\,,
\]
for some $\tilde{p}$ between $p$ and $p^*(\theta)$. Note that by optimality of $p^*(\theta)$, we have $r'(p^*(\theta)) = 0$. Further, \eqref{2-der} implies that
\[
r(p) \le r(p^*(\theta)) - \frac{1}{520}  (p-p^*(\theta))^2\,.
\]

Finally to prove item $(iii)$, taking derivative of both sides of \eqref{optimal-price} with respect to $\theta$ gives us
\[
p^*(\theta) + \theta \frac{\de}{\de \theta} p^*(\theta) = \left(1 - p^*(\theta) - \theta \frac{\de}{\de \theta}p^*(\theta)\right) e^{\theta(1-p^*(\theta))}\,,
\]
and therefore by rearranging the terms
\begin{align*}
\frac{\de}{\de \theta} p^*(\theta) = -\frac{1}{\theta} \left(p^*(\theta) + \frac{e^{\theta(1-p^*(\theta))}}{1+e^{\theta(1-p^*(\theta))}} \right)\,.
\end{align*}
Since $e^{\theta(1-p)} \ge e^{-5/6}$, for $\theta\in {{\sf \Theta}}$ and $p\in \mathcal{P}$, we get
\[
\Big|\frac{\de}{\de \theta} p^*(\theta)\Big| \ge \frac{2}{5}\left(\frac{2}{5} + \frac{e^{-5/6}}{1+e^{-5/6}}\right)\ge 0.2
\]
The result then follows by an application of the Mean Value Theorem.
\subsection{Proof of Lemma~\ref{lem:uncertainty}}\label{proof:lem-uncertainty}
By employing the chain rule for KL divergence~\cite[Theorem 2.5.3]{cover2012elements}, 
\begin{align*}
\KL(f_t^{\pi,\theta_0};f_t^{\pi,\theta}) &= \sum_{s=1}^t \KL(f_t^{\pi,\theta_0};f_t^{\pi,\theta}| \bY_{s-1})\\
&= \sum_{s=1}^t \sum_{\by_s\in \{0,1\}^s} f_s^{\pi,\theta_0}(\by_s) \log\left(\frac{f_s^{\pi,\theta_0}(y_s|\by_{s-1})}{f_s^{\pi,\theta}(y_s|\by_{s-1})}\right)\\ 
&= \sum_{s=1}^t \sum_{\by_{s-1}\in \{0,1\}^{s-1}} f_{s-1}^{\pi,\theta_0}(\by_{s-1}) \sum_{y_s\in \{0,1\}} f_s^{\pi,\theta_0}(y_s|\by_{s-1}) \log\left(\frac{f_s^{\pi,\theta_0}(y_s|\by_{s-1})}{f_s^{\pi,\theta}(y_s|\by_{s-1})}\right)\\ 
&= \sum_{s=1}^t \sum_{\by_{s-1}\in \{0,1\}^{s-1}} f_{s-1}^{\pi,\theta_0}(\by_{s-1}) \KL(f_s^{\pi,\theta_0}(y_s|\by_{s-1}); f_s^{\pi,\theta}(y_s|\by_{s-1}))\\
&\le \sum_{s=1}^t \frac{1}{q(p_s,\theta)(1-q(p_s,\theta))} \sum_{\by_{s-1}\in \{0,1\}^{s-1}} f_{s-1}^{\pi,\theta_0}(\by_{s-1})  (q(p_s;\theta_0)-q(p_s;\theta))^2\\
&\le {7}  \sum_{s=1}^t \sum_{\by_{s-1}\in \{0,1\}^{s-1}} f_{s-1}^{\pi,\theta_0}(\by_{s-1})  (q(p_s;\theta_0)-q(p_s;\theta))^2\,,
\end{align*}
where in the penultimate step we used the inequality $\KL(B_1;B_2)\le \frac{(q_1-q_2)^2}{q_2(1-q_2)}$ for two Bernoulli random variables $B_1$ and $B_2$ with parameters $q_1$ and $q_2$, respectively~\cite[Corollary 3.1]{taneja2004relative}. In the last step we used that $q(p_s,\theta)$ for $p_s\in \mathcal{P}$ and $\theta\in {{\sf \Theta}}$. Next, by using Lemma~\ref{lem:properties} (item 1) we obtain
\begin{align*}
\KL(f_t^{\pi,\theta_0};f_t^{\pi,\theta}) &\le {7} (\theta_0-\theta)^2 \sum_{s=1}^t \sum_{\by_{s-1}\in \{0,1\}^{s-1}} f_{s-1}^{\pi,\theta_0}(\by_{s-1})  (p^*(\theta_0)-p_s)^2\\
& =  {7} (\theta_0-\theta)^2 \sum_{s=1}^t \E_{\theta_0} (p^*(\theta_0)-p_s)^2\,,
\end{align*}
where we used the observation that $p_s$ is a measurable function of $\by_{s-1}$ and $\E_{\theta_0}$ denotes expectation with respect to $f_{s-1}^{\pi,\theta_0}$ measure.
Next by using Lemma \ref{lem:properties} (item 2), we get
\begin{align*}
\KL(f_t^{\pi,\theta_0};f_t^{\pi,\theta}) \le  {3640} (\theta_0-\theta)^2 \sum_{s=1}^t \E_{\theta_0} (r(p^*(\theta_0))-r(p_s))\le {3640} (\theta_0-\theta)^2 \Regret^{\pi}(t,\theta_0)\,.
\end{align*}

\end{document}

%% file: notation.tex
\def\cL{\mathcal{L}}
\def\KL{{\rm KL}}
\def\by{\mathbf{y}}
\def\bY{\mathbf{Y}}

\def\tx{\widetilde{x}}

\def\cF{{\cal F}}

\def\cC{{\cal C}}
\def\cG{{\cal G}}

\def\cG{\mathcal{G}}

\def\reals{{\mathbb R}}

\def\prob{{\mathbb P}}
\def\E{{\mathbb E}}

\def\L0{{L_i}}

\def\de{{\rm d}}
\def\<{\langle}
\def\>{\rangle}
\def\diag{{\rm diag}}

\def\hth{\widehat{\theta}}

\def\hSigma{\widehat{\Sigma}}

\def\F{{\sf F}}

\def\F{{\sf F}}

\def\sT{{\sf T}}

\def\hgamma{\widehat{\gamma}}

\def\v*{v_i}
\def\T*{T_i}

\def\u*{u_i}
\def\F*{F_i}

\def\htheta{\widehat{\theta}}

\def\tx{\tilde{x}}

\def\cL{\mathcal{L}}

\def\hth{{\widehat{\theta}}}

\def\diag{{\rm diag}}

\def\Reg{{\sf Regret}}

\def\l1u{W}

\def\bth0{{\boldsymbol{\theta}_0}}

\def\<{\langle}
\def\>{\rangle}
\def\l1u{W}
\def\rev{{\rm{rev}}}
\def\Regret{{\sf Regret}}
\def\hth{\widehat{{\theta}}}
\def\hpsi{\widehat{{\nu}}}
\def\hgamma{\widehat{{\gamma}}}
\def\mreg{{\sf reg}}